%% file: visionspeech_semantic_2018.tex
\documentclass[journal]{IEEEtran}

\usepackage{graphicx}
\graphicspath{{fig/}}

\usepackage{microtype}
\usepackage{url}
\newcommand{\system}[1]{{\small \textsc{#1}}}
\newcommand{\myparagraph}[1]{\quad{\bf #1}}
\usepackage{latexsym}
\usepackage[cmex10]{amsmath}
\usepackage{amssymb}
\usepackage{wasysym}
\usepackage{cancel}

\renewcommand{\vec}[1]{\boldsymbol{#1}}

\usepackage{booktabs}
\usepackage{multirow}
\usepackage{tabularx}
\newcommand{\mytable}{
    \centering
    \renewcommand{\arraystretch}{1.2}
    }
\newcolumntype{C}{>{\centering\arraybackslash}X}
\newcolumntype{L}{>{\raggedright\arraybackslash}X}
\newcolumntype{R}{>{\raggedleft\arraybackslash}X}

\usepackage{cite}
\newcommand{\citet}[2]{#1~\cite{#2}}

\usepackage[prependcaption,textsize=scriptsize]{todonotes}
\begin{document}

\title{Semantic speech retrieval with a \\ visually grounded model of untranscribed speech}
\author{
    Herman Kamper, Gregory Shakhnarovich, and Karen Livescu
\thanks{H.\ Kamper is with the Department of E\&E Engineering, Stellenbosch University, South Africa  (email: kamperh@sun.ac.za).}%
\thanks{G.\ Shakhnarovich and K.\ Livescu are with TTI-Chicago, USA (email: greg@ttic.edu, klivescu@ttic.edu)}%
}

\markboth{Accepted to the IEEE/ACM Transactions on Audio, Speech, and Language Processing, 2018}
{Kamper, Shakhnarovich and Livescu}
\IEEEpubid{\copyright~2018 IEEE}

\maketitle

\input{abstract}

\IEEEpeerreviewmaketitle

\input{introduction}
\input{related_work_KL}

\input{model}
\input{task}

\input{experiments}
\input{conclusion}

\newpage
\bibliography{mybib}

\end{document}

%% file: abstract.tex
\begin{abstract}
There is growing interest in models that can learn from unlabelled speech paired with visual context.  This setting is relevant for low-resource speech processing, robotics, and human language acquisition research.  Here we study how a visually grounded speech model, trained on images of scenes paired with spoken captions, captures aspects of semantics.  We use an external image tagger to generate soft text labels from images, which serve as targets for a neural model that maps untranscribed speech to (semantic) keyword labels.  We introduce a newly collected data set of human semantic relevance judgements and an associated task, \textit{semantic speech retrieval}, where the goal is to search for spoken utterances that are semantically relevant to a given text query. Without seeing any text, the model trained on parallel speech and images achieves a precision of almost 60\% on its top ten semantic retrievals.  Compared to a supervised model trained on transcriptions, our model matches human judgements better by some measures, especially in retrieving non-verbatim semantic matches.  We perform an extensive analysis of the model and its resulting representations.\footnote{The collected data set of semantic labels is available at:\\ {\url{https://github.com/kamperh/semantic_flickraudio}.}\\The code recipe for the neural networks developed in this work is available at: \url{https://github.com/kamperh/recipe_semantic_flickraudio}.}
\end{abstract}


%% file: introduction.tex
\section{Introduction}
\label{sec:introduction}

\IEEEPARstart{T}{he}
last few years have seen great advances in automatic speech recognition (ASR). 
However, current methods require large amounts of transcribed data, which are available only for a small fraction of the world's languages~\cite{adda+etal_sltu16}.
This has prompted work on speech models that, instead of using exact transcriptions, can learn from weaker forms of supervision, e.g., known word pairs~\cite{synnaeve+etal_slt14,settle+etal_interspeech17}, translation
text~\cite{duong+etal_naacl16,bansal+etal_eacl17,weiss+etal_interspeech17}, or unordered word labels~\cite{palaz+etal_interspeech16}.
The motivation for much of this work is that, even when high-quality ASR is infeasible, it may still be possible to learn low-resource models for practical tasks like retrieval and keyword prediction. 

In this work we consider the setting where untranscribed speech is paired with images.
Such visual context could be used to ground speech when it is not possible to obtain transcriptions, e.g., for endangered or unwritten languages~\cite{chrupala+etal_arxiv17}.
In robotics, co-occurring audio and visual signals could be combined to learn new commands~\cite{luo+etal_icvs08,krunic+etal_icra09,taniguchi+etal_advrob16}.

Our work builds on a line of recent studies~\cite{synnaeve+etal_nipsworkshop14,harwath+etal_nips16,chrupala+etal_acl17,drexler+glass_glu17,leidal+etal_asru17,scharenborg+etal_arxiv18,harwath+etal_arxiv18,harwath+etal_icassp18} that use natural images of scenes paired with spoken descriptions.
Neither the spoken nor visual input is labelled.
Most approaches map the images and speech into some common space ($\S$\ref{sec:background_vision+speech}), allowing images to be retrieved using speech and vice versa.
Although useful, such models cannot predict (written) labels for the input speech.

Here we present an expansion of our work in~\cite{kamper+etal_interspeech17}, 
where we proposed a model that can make such text label predictions.
A trained visual tagger is used to obtain soft text labels for each training image, and these are used as targets for a neural network that maps speech to keyword labels. 
The resulting model can be used to predict which (written) words are present in an unseen
spoken utterance. 
Without observing any transcriptions, 
the model can be used as a keyword spotter, predicting which utterances in a search collection contain a given written keyword.
In an initial qualitative analysis, we observed that the model often confuses semantically related words 
such as `man' and `person';
these count as errors in keyword spotting, but should be
useful in {\it semantic} search applications.

Our primary aim here is to expand our work in~\cite{kamper+etal_interspeech17} by performing an extensive analysis 
to see
what aspects of semantics are captured by the model. 
To do so formally, we use the task of \textit{semantic speech retrieval}, where the aim is to retrieve all utterances in a speech collection that are semantically relevant
to a given query keyword, irrespective of whether that keyword occurs exactly in an utterance or not.
E.g., given the query `children', the goal is to return not only utterances containing the word `children', but also utterances about children, like `young boys playing soccer in the park'.
There has been some work on this and related tasks ($\S$\ref{sec:background_speech_retrieval}), but {typically in higher-resource settings} 
and none using visual supervision.
{Note that visual supervision could actually prove more useful than transcriptions for semantic retrieval in some cases (and we make findings in this direction).} 
{To our knowledge, despite the prior work, speech data with soft semantic relevance judgements does not exist. This limits analyses of semantics, which are often ambiguous.} 
We therefore 
collect and release
a new data set {with semantic labels from multiple annotators}.
Using this data, we present an extensive analysis of an updated version of the model of~\cite{kamper+etal_interspeech17}, and compare it to several new alternative models for the task of semantic speech retrieval.

\enlargethispage{-1.2\baselineskip} 

Our main contributions are to extensively analyse the visually grounded speech modeling approach in the context of a formal semantic task:  We provide the new data set, compare model performance to human judgements, study the learned representations, and analyse the performance of the model against multiple alternatives.
We find
that the predictions from the model match well with human
judgements, leading to competitive retrieval scores, particularly in retrieving 
non-exact semantic matches.
Specifically,
the model even outperforms a supervised model trained on transcriptions when searching for utterances that are semantic but not verbatim matches to the keyword.
We conclude that visual context can play an important role in capturing semantics in speech, especially but not only in the absence of textual supervision.

%

%% file: related_work_KL.tex
\section{Related Work}
\label{sec:related}

\subsection{Language acquisition}
\label{sec:background_language_acquisition}

Cognitive scientists have long been interested in how infants use sensory input to learn the mapping of words to real-world entities~\cite{yu+smith_psych07,cunillera+etal_qjep10,thiessen_cogsci10}, with computational models providing one way to specify and test theories.
\citet{Roy and Pentland}{roy+pentland_cogsci02} and \citet{Yu and Ballard}{yu+ballard_tap04} were some of the first to consider real speech input, followed by more recent work such as~\cite{tenbosch+cranen_interspeech07,aimetti_eacl09,driesen+vanhamme_neurocomp11}.
However, these studies simplify the problem by using discrete labels to represent the visual context~\cite{rasanen+rasilo_psych15}, the spoken input~\cite{gelderloos+chrupala_coling16}, or both~\cite{siskind_cognition96,frank+etal_psych09}; infants do not have access to such idealised input.
Our model operates on natural images paired with real unlabelled speech.
One assumption we make is that a visual tagger is available for processing training images.
Our focus is not on cognitive modelling, but this assumption is linked to the question of whether visual category acquisition precedes word learning in infants~\cite{clark_trendscogsci04}.
We leave the cognitive implications of our model for future work.

Also motivated by the problem of modelling language acquisition, there is growing interest in the speech community in models that can learn directly from unlabelled speech~\cite{jansen+etal_icassp13,versteegh+etal_sltu16}.
These studies have the additional motivation that such systems could be applied in settings where it is difficult or impossible to obtain transcriptions.
The goal here is to learn exclusively from audio, i.e.\ without any supervision signal, even from another modality.
Tasks include acoustic unit discovery~\cite{varadarajan+etal_acl08,lee+glass_acl12}, query-by-example search~\cite{zhang+etal_icassp12}, unsupervised term discovery~\cite{park+glass_taslp08}, and unsupervised segmentation and clustering~\cite{lee+etal_tacl15,kamper+etal_taslp16}. See~\cite{kamper_phd16} for a complete literature review.
Although our work is
related to these studies, we focus on the setting where visual context is available for grounding.

\subsection{Joint modelling of images and text}
\label{sec:background_vision+text}

Joint modelling of images and text has received much recent attention.
One approach is to map images and text into a common
vector space where related instances are close to each other, e.g., for text-based image retrieval~\cite{socher+feifei_cvpr10,weston+etal_ijcai11,hodosh+etal_jair13,karpathy+etal_nips14}.
Image captioning has also been studied extensively, where the goal is to produce a natural language description for a visual scene~\cite{farhadi+etal_eccv10,yang+etal_emnlp11,kulkarni+etal_pami13,bernardi+etal_jair16}. 
Most recent approaches use a convolutional neural network (CNN) to
convert an input image to a latent representation, which is then fed
to a recurrent neural network to produce a sequence of output
words~\cite{donahue+etal_cvpr15,fang+etal_cvpr15,karpathy+feifei_cvpr15,vinyals+etal_cvpr15,xu+etal_icml15,chen+lawrencezitnick_cvpr15}. 
Our work uses spoken rather than written language, and we consider a semantic speech retrieval task.

There has also been work on using vision to
explicitly
capture aspects of text semantics.
Semantics are difficult to annotate,
so most studies
evaluate models
using soft human ratings on tasks such as word similarity~\cite{feng+lapata_naacl10,silberer+lapata_emnlp12,bruni+etal_jair14}, word association~\cite{bruni+etal_acl12,roller+imwalde_emnlp13} or concept categorisation~\cite{silberer+lapata_acl14}.
We also use human responses,
but for the task of semantic speech retrieval. 

\subsection{Joint modelling of vision and speech}
\label{sec:background_vision+speech}

Recent work has shown that ASR can be improved by including additional visual features from the scene in which speech occurs~\cite{sun+etal_slt16,gupta+etal_icassp17}.
These studies consider fully supervised ASR, and typically the scene is not described by the speech but is complementary to it.
Our aim instead is to use the visual modality to learn from matching but \textit{untranscribed} speech.

This is the setting considered by~\citet{Synnaeve et al.}{synnaeve+etal_nipsworkshop14} and \citet{Harwath et al.}{harwath+etal_nips16}, who used natural images of scenes paired with unlabelled spoken descriptions to learn neural mappings of images and speech into a common space.
This approach allows images to be retrieved using speech and vice versa. This is useful, e.g., in applications for tagging mobile phone images with spoken descriptions~\cite{hazen+etal_interspeech07,anguera+etal_icmir08}.
The joint neural mapping approach has subsequently been used for spoken word and object segmentation~\cite{harwath+glass_acl17}, and the learned representations have been analysed in a variety of ways~\cite{chrupala+etal_arxiv17,drexler+glass_glu17}.
Despite these developments, this prior work does not give an explicit mapping of speech to textual labels.

The model we proposed in~\cite{kamper+etal_interspeech17}
can make such labelled predictions: a visual tagger is used to obtain soft labels for a given training image, and a neural network is then trained to map input speech to these targets.
The resulting speech model attempts to predict which (written) words are present in a spoken utterance, ignoring the order and quantity of words.
The model can be used as a keyword spotter, retrieving utterances in a speech collection that contain a given keyword. It was found that the model also retrieves semantic matches for a query, not only exact matches.  However, we did not formalise the semantic search task, compare results to human judgements, or extensively analyse the performance of the model against multiple other systems.
Here we present an extensive analysis using several alternative models for semantic speech retrieval.

\subsection{Semantic text retrieval}
\label{sec:background_text_retrieval}

In textual information retrieval, the task is to find text documents in a collection that are relevant to a given written keyword, irrespective of whether the keyword occurs exactly in the document.
This can be accomplished using \textit{query expansion},
where additional words or phrases similar in meaning to the query are used to match relevant documents~\cite{xu+croft_sigir96,graupmann+etal_vldb05}.
Classic approaches use co-occurrences of words, or resources such as WordNet~\cite{miller_acm95}, to expand the query list, while recent methods use word embeddings~\cite{diaz+etal_acl16,roy+etal_sigir16}. 
We also consider a semantic search task, but on speech rather than text.  We do use text-based methods to obtain an upper bound on performance using the transcriptions of the speech collection ($\S$\ref{sec:baseline_oracle}).

\subsection{Keyword spotting and (semantic) speech search}
\label{sec:background_speech_retrieval}

In the speech research literature, keyword spotting and spoken term detection involve
retrieving utterances in a search collection that contain exact spoken instances of a given written keyword~\cite{wilpon+etal_assp90,szoke+etal_interspeech05,garcia+gish_icassp06}.
Here we specifically consider \textit{semantic} rather than exact retrieval.
There has been some work on semantic speech retrieval~\cite{garofolo2000trec,chelba+etal_ieee08,lee+etal_slt12,lee+etal_taslp15}; the most common approach is to cascade ASR with text-based retrieval.
This, again, requires large amounts of labelled data with which to train an accurate ASR system.
There have been some attempts to do semantic retrieval in the absence of transcriptions by extending unsupervised speech modelling approaches (described in $\S$\ref{sec:background_language_acquisition}). In~\cite{li+etal_asru13}, an unsupervised term detection system was used to automatically discover reoccurring words or phrases. Given a spoken query, the system then expanded the query using other commonly co-occurring discovered terms, thereby allowing retrieval of semantically related words (other than the one spoken). In~\cite{harwath+etal_icassp13}, a topic model was applied to discovered words in a similar way.
Here we are also doing semantic speech retrieval in the absence of transcriptions. {Our task is different, however, in that our queries are textual keywords; this is not possible using a model trained without some grounding signal (which is why the above studies performed query-by-example search).} 
We are considering the case where we have limited supervision in the form of visual context. 
This is a first, to our knowledge.

%% file: model.tex
\section{The Visually Grounded Speech Model}
\label{sec:model}

Given a corpus of parallel images and spoken captions, neither with textual labels, we train a spoken keyword prediction model using a visual tagging system to produce soft labels for the speech network.
Originally, we used the model for keyword spotting~\cite{kamper+etal_interspeech17}, with the output of the model used to predict whether a given written keyword occurs exactly in a test utterance.
Here, we also interpret the output of the model as a score for a keyword, but instead of looking for an exact match, we use it for predicting semantic relevance. 

The overall approach is illustrated in Figure~\ref{fig:vision_speech_cnn}. 
Training image $I$ is paired with a spoken caption 
$X = \vec{x}_1, \vec{x}_2, \ldots, \vec{x}_T$, where each frame $\vec{x}_t$ is an acoustic feature vector, e.g.\ Mel-frequency cepstral coefficients (MFCCs)~\cite{davis+mermelstein_tassp80}.
We use an external vision system to tag $I$ with soft textual labels, giving targets $\hat{\vec{y}}_\textrm{vis}$ to train the speech network $\vec{f}(X)$. 
The resulting network $\vec{f}(X)$ can then predict which keywords are present for a given utterance $X$, disregarding the order, quantity, or locations of the keywords in the input speech.
The possible
keywords (the vocabulary) are implicitly specified by the visual tagger, and no transcriptions are used during training.
When applying the trained $\vec{f}(X)$ for keyword spotting or semantic retrieval, test speech alone is used without any
visual input.

\subsection{Detailed model description}
\label{sec:model_details}

\begin{figure}[t]
    \centering
        \includegraphics[width=0.975\linewidth]{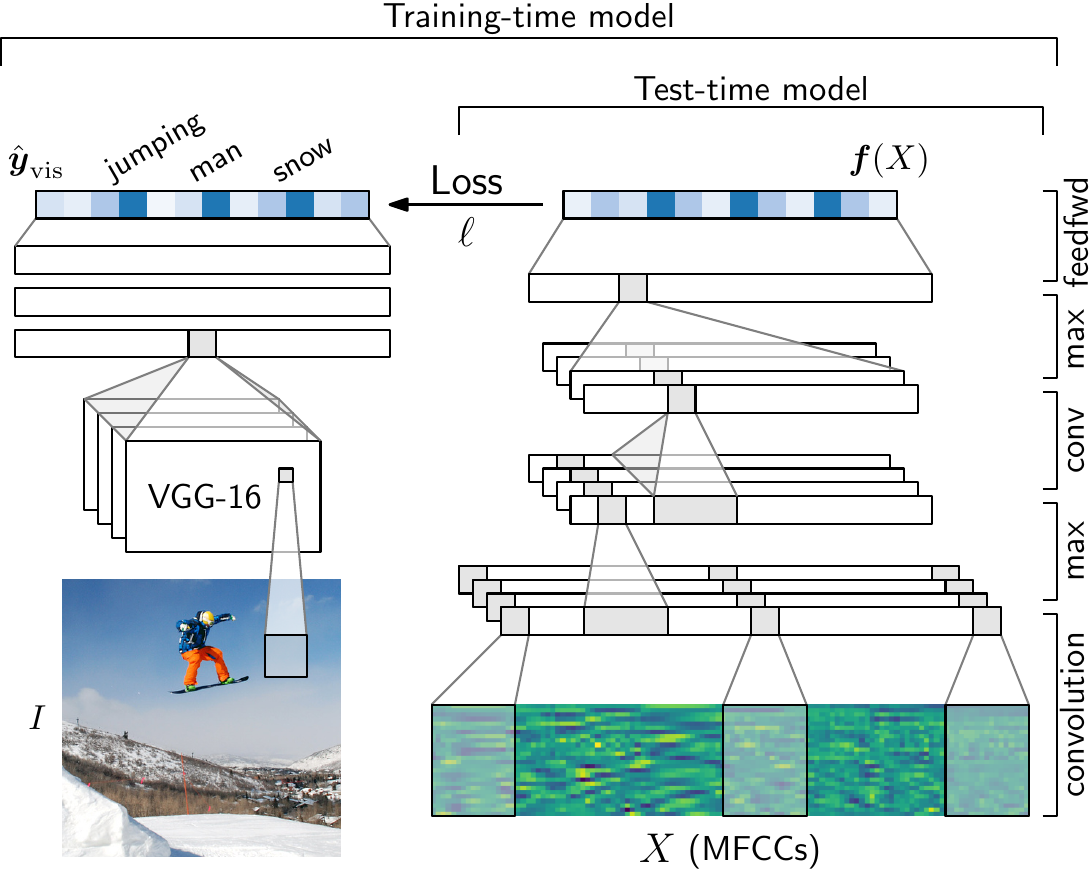}
    \caption{For training, an external visual tagger produces soft tags for image $I$, which serve as targets for the speech network fed with spoken caption $X$. In testing, the speech network is given unseen speech (no image) and the output $f(X)$ is used for semantic retrieval of a textual keyword. 
    }
    \label{fig:vision_speech_cnn}
\end{figure}

{In typical speech applications like keyword spotting, supervision consists of text labels of the keywords that appear in the speech: for }
training utterance $X$, we could construct a multi-hot vector $\vec{y}_{\textrm{bow}} \in \{0,1\}^W$, with $W$ the vocabulary size and each dimension $y_{\textrm{bow}, w}$ a binary indicator for whether word $w$ occurs in $X$.
The order and quantity of words are ignored in this type of bag-of-words (BoW) labelling~\cite{palaz+etal_interspeech16}. 
Instead of a transcription for $X$, we only have 
the paired image $I$.
We use a multi-label visual tagging system which, instead of binary indicators, produces soft targets $\hat{\vec{y}}_{\textrm{vis}} \in [0,1]^W$, with $\hat{y}_{\textrm{vis}, w} = P_{\vec{\gamma}}(w | I)$ the estimated probability of word $w$ being present in image $I$ under vision model parameters $\vec{\gamma}$.
In Figure~\ref{fig:vision_speech_cnn}, $\hat{\vec{y}}_{\textrm{vis}}$ would ideally be close to $1$ for $w$ corresponding to words such as
`jumping', `man', `snow' and `snowboard', and close to $0$ for irrelevant dimensions.
{For semantic retrieval, we therefore interpret $\hat{\vec{y}}_{\textrm{vis}}$ as a bag-of-visual (semantic) tags.}
This visual tagger is fixed:
during training of the speech network $\vec{f}(X)$, as described below, vision parameters $\vec{\gamma}$ are never updated.

Given $\hat{\vec{y}}_{\textrm{vis}}$ as target, we train the speech
model $\vec{f}(X)$ (Figure~\ref{fig:vision_speech_cnn}, right).
This model (parameters $\vec{\theta}$) consists of a convolutional neural network (CNN) over the speech $X$ with a final sigmoidal layer so that $\vec{f}(X) \in [0,1]^W$.
We interpret each dimension of the output as $f_w(X) = P_{\vec{\theta}}(w | X)$.
Note that $\vec{f}(X)$ is not a distribution over the output
vocabulary, since any number of keywords can be present in an
utterance; rather, it is a multi-label classifier where each dimension
$f_w(X)$ can have any value in $[0,1]$.  
We train the speech
model using the summed cross-entropy loss, which (for a single training example) is:
\begin{align}
    \ell(\vec{f}(X), \hat{\vec{y}}_\textrm{vis}) 
    &= -\sum_{w = 1}^W \left\{ \hat{y}_{\textrm{vis}, w} \log f_w(X) \;\; +\right. \nonumber \\ 
    &\left. (1 - \hat{y}_{\textrm{vis}, w}) \log\left[1 - f_w(X) \right] \right\}  
    \label{eq:summed_cross_entropy}
\end{align}
If we had $\hat{y}_{\textrm{vis}, w} \in \{0, 1\}$, as in $\vec{y}_\textrm{bow}$, this
would be the summed log loss of $W$
binary classifiers.
A  pre-trained visual network was also used to provide supervision for another modality using a similar loss in~\cite{aytar+etal_nips16}, where video was paired with general audio (not speech).

\subsection{The visual tagging system}
\label{sec:vision_system}

In image classification, the task is to
choose one (object) class from
a closed set~\cite{deng+etal_cvpr09}, while in image captioning, the goal is to produce a natural language description
of a scene~($\S$\ref{sec:background_vision+text}).
In contrast to both these tasks, 
we require a visual tagging system~\cite{barnard+etal_jmlr03,guillaumin+etal_iccv09,chen+etal_icml13} that predicts an unordered set of words (nouns, adjectives, verbs) that accurately describe aspects of the scene (Figure~\ref{fig:vision_speech_cnn}, left).
This is a multi-label binary classification task.

We train our visual tagger on
data from the
Flickr30k~\cite{young+etal_tacl14} and MSCOCO~\cite{lin+etal_eccv14} data sets, which consist of images
each
with five written captions.
We combine Flickr30k with the training portion of MSCOCO, and
remove any images that occur in the parallel image-speech data used in our experiments ($\S$\ref{sec:data}). 
The result is a training set of around 106k images, significantly more than the 25k
used in~\cite{kamper+etal_interspeech17}.
For each image, a single target BoW vector $\vec{y}_\textrm{vis}$
is constructed by combining all five captions
after removing stop words.
Element ${y}_{\textrm{vis}, w}$ is an indicator for whether word $w$ occurs in any of the five captions
for that image, where $w$ is one of the $W = 1000$ most common content words in the combined set of image captions.

We follow the
common practice of using a pre-trained visual representation for processing the images (Figure~\ref{fig:vision_speech_cnn}, left).
Specifically, we use VGG-16~\cite{simonyan+zisserman_arxiv14}, trained on around 1.3M images~\cite{deng+etal_cvpr09}; we replace the final classification layer with four 2048-unit {rectified linear unit (ReLU)} layers, followed by a final sigmoidal layer for predicting word occurrence.
The visual tagger, with parameters $\vec{\gamma}$, is then trained on the combined Flickr30k and MSCOCO data 
using the summed log loss~\eqref{eq:summed_cross_entropy} with arguments $\ell\left(\hat{\vec{y}}_\textrm{vis}, {\vec{y}}_\textrm{vis}\right)$.
The VGG-16 parameters are fixed;
only the additional fully connected layers
are updated.

Previous vision-speech models~\cite{synnaeve+etal_nipsworkshop14,harwath+etal_nips16,chrupala+etal_acl17}
also employ pre-trained visual representations. 
Here we take this approach even further by using the textual classification output of a trained vision system.
Although we train (and then fix) the visual tagger ourselves, we ensure that none of the training data overlaps with the parallel image-speech data used in our speech model, so the model does not obtain even indirect access to text~labels.

%% file: task.tex
\section{Semantic Speech Retrieval}
\label{sec:task}

We are interested in whether our model can be used to determine the semantic concepts present in a speech utterance.  I.e., can we use the model to search a speech collection
for mentions of a particular semantic concept?  We formalise this task and collect a new data set for evaluation.

\subsection{Task description}
\label{sec:task_description}

{We consider the task of \textit{semantic speech retrieval}.}
Instead of matching keywords exactly, as is typical in keyword spotting ($\S$\ref{sec:background_speech_retrieval}), the aim is to retrieve all utterances that 
are semantically relevant, irrespective of whether the keyword occurs in the utterance or not.
E.g., for the query `sidewalk', a model should return not only utterances containing the word exactly, 
but also speech like `an old couple window-shopping on a Paris street.'
As noted, there has been some work on semantic speech retrieval, but typically in higher-resource settings and none using visual supervision ($\S$\ref{sec:background_speech_retrieval}).
Furthermore, the new data set described below contains labels from multiple annotators, allowing for a fine-grained analysis.



\subsection{Data set}
\label{sec:semantic_data}

\begin{figure}[b]
    \centering
    \begingroup
    \small
    \def\arraystretch{1.5}
    \begin{tabularx}{1\linewidth}{|L|}
        \hline
        For the sentence below, select all words that could be used to search for the described scene.
        
        \vspace{5pt}
        
        \textbf{Sentence:} a skateboarder in a light green shirt.
        
        \vspace{5pt}
        
        $\Square$~dogs \ $\Square$~beach \ $\Square$~children \ $\Square$~white \ $\Square$~swimming \ $\Square$~wearing \ $\Square$~skateboard \ $\Square$~None of the above
        \\
        \hline
    \end{tabularx}
    \endgroup
    \caption{An example AMT job for semantically annotating a transcription of a spoken sentence with keywords. In this case, `skateboard' was selected by all five annotators and `wearing' by~four. 
    }
    \label{fig:turk}
\end{figure}

We 
extend the corpus of \cite{harwath+glass_asru15}, which consists of parallel images and spoken captions. 
The data is transcribed, but not semantically labelled. 
For a subset of the speech in the corpus, we use Amazon Mechanical Turk (AMT) to collect semantic labels from human annotators.

As our keywords, we select a set of 70 random words from the transcriptions of the training portion of the corpus, ignoring stop words.  
The test portion of the spoken caption data consists of 1000 images, each with five spoken captions; we collect semantic labels for one randomly selected spoken caption from each of these 1000 images.
A single AMT job consists of the transcription of a single utterance (describing a scene) with a list of seven potential keywords from which an annotator could select any number, as illustrated in Figure~\ref{fig:turk}.
To cover all 70 keywords, a given sentence is repeated over ten jobs.
Since the question of semantic relatedness (between a given sentence and keyword) is inherently ambiguous, we have five
workers annotate each utterance.
Annotators were not shown any images but only the utterance transcriptions and candidate keywords, so the task is the same as in other work on semantic speech retrieval  ($\S$\ref{sec:background_speech_retrieval}).

To analyse annotator agreement, we consider the number of annotators (between zero and five) that selected a given keyword for each sentence.
Table~\ref{tbl:turk_counts} gives the proportion of (sentence, keyword) pairs selected by a given number of annotators. 
Annotators agree most often (83.3\% of the time) about the absence of a keyword for a given sentence, 
while there are very few examples 
that all annotators agree are semantically matched (1.6\%).

To evaluate a semantic keyword retrieval model against the human annotations, one option is to combine the human judgements into a single hard label.
On the other hand, the fact that there is a wide range of opinions among the human annotators indicates that semantic relevance may be inherently ``soft'', motivating evaluation by comparing against the proportion of annotators that agree with a given label.
In work using vision to explicitly aspects of semantics in {text}, soft human ratings were also used (\S\ref{sec:background_vision+text}).
In our experiments we consider both hard and soft options.

\begin{table}[t]
    \mytable
    \caption{The proportion of (sentence, keyword) pairings selected by a given number of annotators.}
    \begin{tabularx}{\linewidth}{@{}lCCCCCC}
    \toprule
    Count       & 0 & 1 & 2 & 3 & 4 & 5\\
    Proportion (\%)  & 83.3 & 6.9 & 3.0 & 2.6 & 2.5 & 1.6 \\
    \bottomrule
    \end{tabularx}
    \label{tbl:turk_counts}
\end{table}

To obtain a hard label of whether a {sentence} is semantically relevant or irrelevant {given a keyword},
we take the majority decision: if three or more annotators selected that keyword, we label that {sentence} 
as relevant for that {keyword}
, otherwise we label it as irrelevant.
{Under this assignment, 
95.8\% of annotators} agree with the hard decision (bearing in mind the skew towards negative assignments).
{We removed three keywords with very poor agreement, leaving 67 keywords.} 

The result of our data collection effort is
a data set of 1000 spoken utterances, each annotated with a set of keywords that could be used to search for that utterance.

%% file: experiments.tex
\section{Experimental Setup and Evaluation}
\label{sec:data}

We train our model on the corpus of parallel images and spoken captions of~\cite{harwath+glass_asru15}, containing 8000 images with five spoken captions each.
The audio comprises around 37~hours of active speech. 
The data comes with train, development and test splits containing 30\,000, 5000 and 5000 utterances, respectively. 
As described above, we obtained semantic labels for 1000 of the test utterances.
We parameterise the speech audio as 13 MFCCs with first and second order derivatives, giving 39-dimensional input vectors.\footnote{We also tried filterbanks; MFCCs always gave similar or slightly better performance on development data.}
99.5\% of the utterances are 8~s or shorter; utterances longer than this are truncated while utterances shorter than this are zero-padded to 8~s (800 frames).
We deal with the variable duration of utterances by pooling all units at the output of convolutional layers over time (see below).\footnote{
We also experimented with using the actual durations of the utterances, but found minimal difference.
Using the actual durations incurs an additional computational cost, whereas our approach can ignore the zero-padded regions through the global max pooling operation over time.
}

Training images are passed through the visual tagger ($\S$\ref{sec:vision_system}), producing soft targets $\hat{\vec{y}}_\textrm{vis}$ for training the keyword prediction model $\vec{f}(X)$ on the unlabelled speech ($\S$\ref{sec:model_details}), as shown in Figure~\ref{fig:vision_speech_cnn}.
We refer to the resulting model as \system{VisionSpeechCNN}, which during testing is presented only with spoken input. 
We are interested in \textit{semantic} speech retrieval ($\S$\ref{sec:task}), but labels for this task are only available for test utterances.
We therefore optimise the hyper-parameters of the model using \textit{exact} keyword spotting on development data {using the same set of 67 keywords; each spoken caption in the development data is treated as a separate test item}.
\system{VisionSpeechCNN} has the following resulting structure:
1-D ReLU convolution with 64 filters over 9 frames; max pooling over 3 units; 1-D ReLU convolution with 256 filters over 10 units; max pooling over 3 units; 1-D ReLU convolution with 1024 filters over 11 units; max pooling over all units; 3000-unit fully-connected ReLU; and the 1000-unit sigmoid output.
A stride of one unit is used for all convolution operations and there is no overlap between windows in the max pooling operations.
Based on experiments on development data, we train for a maximum of 25 epochs with early stopping using Adam optimization~\cite{kingma+ba_iclr15} with a learning rate of $1\cdot10^{-4}$ and a batch size of eight.\footnote{The code recipe for the neural networks developed in this work is available at:  \url{https://github.com/kamperh/recipe_semantic_flickraudio}.}

\subsection{Evaluation}

To use \system{VisionSpeechCNN} for semantic retrieval, we use its output $f_w(X) \in [0, 1]$ as a score for how relevant an utterance $X$ is given the keyword $w$. 
The baseline and cheating models (described below) similarly predict a relevance score for each utterance given a specific keyword.

We compare a model's predictions to semantic labels obtained from human annotators ($\S$\ref{sec:semantic_data}) using several metrics.
To obtain a hard labelling from a model, we set a threshold $\alpha$, and label all keywords for which $f_w(X) > \alpha$ as relevant.
By comparing this to the ground truth semantic labels (according to majority annotator agreement), precision and recall can be calculated; to measure performance independent of $\alpha$, we report \textbf{average precision (AP)}, the area under the precision-recall curve as $\alpha$ is varied.
Instead of using the hard ground truth semantic labels, the soft scores $f_w(X)$ can also be compared directly to the number of annotators that selected the keyword $w$ for utterance $X$: we use \textbf{Spearman's $\rho$} to measure the correlation between the rankings of these two variables, as is common in work on word similarity~\cite{agirre+etal_naacl09,hill+etal_cl15}.
The remaining metrics are standard in (exact) keyword spotting, based on how a model ranks utterances in the test data from most to least relevant for each keyword~\cite{hazen+etal_asru09,zhang+glass_asru09}: \textbf{precision at ten} ($P@10$) is the average precision of the ten highest-scoring proposals; \textbf{precision at $N$} ($P@N$) is the average precision of the top $N$ proposals, with $N$ the number of true occurrences of the keyword; and \textbf{equal error rate (EER)} is the average error rate at which false acceptance and rejection rates are equal.

Apart from Spearman's $\rho$, all these metrics can also be used to evaluate exact keyword spotting.

\subsection{Baselines and cheating models}
\label{sec:baseline_oracle}

We consider a number of baselines as well as ``cheating'' models which use idealised information not available to \system{VisionSpeechCNN}.

\myparagraph{{Prior}-based baselines:}
\system{TextPrior} uses the unigram probability of each keyword estimated from the transcriptions of the training portion of the spoken captions corpus.
This will indicate how much better our model does than
simply
hypothesising common words.
Similarly, \system{VisionTagPrior} is obtained by passing all training images 
through the trained visual tagger ($\S$\ref{sec:vision_system}), and then taking the average over all images.
It therefore always predicts common visual labels.

\myparagraph{\system{VisionCNN}:}
\system{VisionSpeechCNN} is trained to predict soft visual tags. 
One question is whether it therefore learns to ignore any aspect of the acoustics that does not contribute to predicting the visual target.
The model
\system{VisionCNN} is
an attempt to test this: as the representation for each test utterance, it passes through the visual tagger the \textit{true} image paired with that utterance.
Since it uses ideal information, 
it is
our first cheating model.
If \system{VisionSpeechCNN} were to perfectly predict image tags (ignoring acoustics that do not contribute to visual prediction), then \system{VisionCNN} would be an upper bound on performance.  But in reality our model could do better or worse than \system{VisionCNN}, since the speech contains some information not in the images and training does not generalise perfectly.

\myparagraph{\system{SupervisedBoWCNN}:}
Instead of soft targets from an image tagger, the \system{SupervisedBoWCNN} cheating model uses transcriptions to obtain 
hard BoW supervision ($\S$\ref{sec:model_details}): $\vec{y}_{\textrm{bow}}$ targets are constructed for the 1000 most common words in the transcriptions of the 30\,000 speech training utterances (ignoring stop words) and the loss~\eqref{eq:summed_cross_entropy} is used for training, $\ell(\vec{f}_\textrm{bow}(X), \vec{y}_\textrm{bow})$. Other than ideal supervision, the model has the same structure and training procedure as \system{VisionSpeechCNN}.  This model allows us to explore how the visual supervision differs from typical textual transcriptions for semantic concept learning. \label{sec:oracle}

\myparagraph{Text-based cheating models:}
Suppose we had a perfect ASR system, converting input speech to text without errors.  How well could we do at semantic speech retrieval using this error-free text?
To answer how this cascaded approach would do in an ideal setting, we consider two  text-based semantic retrieval methods ($\S$\ref{sec:background_text_retrieval}) 
using transcriptions of the speech.
The first is based on WuP similarity, named after~\citet{Wu and Palmer}{wu+palmer_acl94}, which scores the semantic relatedness between two words according to the path length between them in the WordNet lexical hierarchy~\cite{miller_acm95}.
For our task, the \system{Text\-WuP} model is based on the closest match (in WuP) between a keyword and each of the
words in a transcribed utterance.
Our second text-based cheating method 
is based on word embeddings.
Specifically, we use the \system{ParagramXXL} embedding method of~\cite{wieting+etal_tacl15,wieting+etal_iclr16}, 
developed for semantic sentence similarity prediction.
For our task, the \system{TextParagram} 
cheating model calculates the cosine similarity between a keyword embedding and the \system{Paragram} sentence embedding of an utterance.
\begin{table}[!t]
    \mytable
    \caption{The retrieved utterance rated highest by \textsc{VisionSpeechCNN} for a selection of keywords. The number of annotators (out of five) that selected the keyword for that utterance is shown, with $*$ indicating an incorrect semantic retrieval according to the majority labelling.} 
    \begin{tabularx}{\linewidth}{@{}lL@{\ }l@{}}
        \toprule
        Keyword & Top retrieved utterance & Count \\
        \midrule
        jumps & biker jumps off of ramp & 5 / 5 \\
        ocean & man falling off a blue surfboard in the ocean & 5 / 5 \\
        race & a red and white race car racing on a dirt racetrack & 5 / 5 \\
        snowy & a skier catches air over the snow & 5 / 5\\
        bike & a dirt biker rides through some trees & 4 / 5 \\
        children & a group of young boys playing soccer & 4 / 5 \\
        riding & a surfer rides the waves & 4 / 5 \\
        young & a little girl in a swing laughs & 4 / 5 \\
        field & two white dogs running in the grass together & 3 / 5 \\
        swimming & a woman holding a young boy slide down a water slide into a pool & 3 / 5 \\
        carrying & small dog running in the grass with a toy in its mouth  & 2 / 5 $*$ \\
        face & a man in a gray shirt climbs a large rock wall & 2 / 5 $*$ \\
        large & a group of people on a zig path through the mountains & 1 / 5 $*$ \\
        sitting & a baby eats and has food on his face & 1 / 5 $*$ \\
        hair & two women and a man smile for the camera & 0 / 5 $*$ \\
        \bottomrule
    \end{tabularx}
    \label{tbl:keyword_spotting_dev_examples}
\end{table}

\begin{table*}[!t]
    \mytable
    \caption{Keyword spotting and semantic speech retrieval performance for \textsc{VisionSpeechCNN}~(row 3), compared against the baseline (rows 1 and 2) and cheating~(rows 4 to 7) models. Boldface indicates both the top-scoring 
    non-cheating model (rows 1 to 3) as well as the best cheating model (rows 4 to 7) for each of the metrics.}
        \begin{tabularx}{\linewidth}{@{}lRRRRcRRRRr}
    \toprule
    & \multicolumn{4}{c}{Exact keyword spotting (\%)} & ~~~ & \multicolumn{5}{c}{Semantic speech retrieval (\%)} \\
    \cmidrule{2-5} \cmidrule(l){7-11}
    Model & $P@10$ & $P@N$ & EER & AP & & $P@10$ & $P@N$ & EER & AP & Spear.\ $\rho$ \\
    \midrule
    {\bf Baseline models:} & & & & & & & & & & \\
    1. \textsc{TextPrior} & 2.8 & 3.4 & 50.0 & 8.7 & & 6.1 & 7.0 & 50.0 & 11.4 & 10.8 \\
    2. \textsc{VisionTagPrior} & 2.8 & 3.4 & 50.0 & 7.0 & & 6.1 & 7.0 & 50.0 & 13.6 & 12.5 \\
    \midrule
    \addlinespace
    3. \textsc{VisionSpeechCNN} & \textbf{38.5} & \textbf{30.8} & \textbf{19.6} & \textbf{26.9} & & \textbf{58.8} & \textbf{39.7} & \textbf{23.9} & \textbf{39.4} & \textbf{32.4} \\
    \addlinespace
    \midrule
    {\bf Cheating models:} & & & & & & & & & & \\
    4. \textsc{VisionCNN} & 31.0 & 26.2 & 22.1 & 22.2 & & 54.2 & 38.9 & {22.8} & 37.4 & \textbf{33.8} \\
    5. \textsc{SupervisedBoWCNN} & \textbf{84.9} & \textbf{74.7} & 5.6 & \textbf{87.3} & & 88.1 & 50.3 & 23.8 & 51.3 & 21.9 \\
    6. \textsc{TextWuP} & 65.4 & 67.3 & \textbf{2.6} & 75.2 & & 80.3 & 63.0 & 19.4 & \textbf{60.9} & 25.2 \\
    7. \textsc{TextParagram} & 80.0 & 72.1 & 3.5 & 67.7 & & \textbf{88.8} & \textbf{64.0} & \textbf{14.3} & 60.1 & 31.6 \\
    \bottomrule
    \end{tabularx}
    \label{tbl:results}
\end{table*}

\section{Experimental Results and Analysis}
\label{sec:experiments}

For a qualitative view, Table~\ref{tbl:keyword_spotting_dev_examples} shows the top retrievals when we use \system{VisionSpeechCNN} to do semantic speech retrieval for a selection of keywords.
The number of annotators that marked the utterance as relevant 
is also shown; $*$ indicates incorrect retrievals according to the ground truth (i.e.\ majority) semantic labelling.
Out of the 15 results shown, ten retrievals are correct.

The quantitative metrics for exact and semantic keyword retrieval for \system{VisionSpeechCNN} and all the baseline and cheating models are shown in Table~\ref{tbl:results}.
In both exact and semantic retrieval, \system{VisionSpeechCNN} 
outperforms the baseline models 
across all metrics.
The baseline models, \system{VisionSpeechCNN}, and \system{VisionCNN} all perform better at semantic than at exact retrieval.
In contrast, the transcription-based cheating models (rows 5 to 7) perform better on $P@10$, but worse on all other semantic search metrics.
$P@10$ only measures precision of the highest ranked utterances, while the other metrics combine precision and recall; thus, the transcription-based cheating models struggle to retrieve semantic matches compared to exact matches, while \system{VisionSpeechCNN} and \system{VisionCNN} recall more semantic matches.
In terms of absolute performance, the transcription-based models 
still perform better at semantic speech retrieval on the metrics based on hard ground truth labels.
However, for Spearman's $\rho$, which gives credit even if a prediction does not match the majority of annotations, \system{VisionCNN} outperforms all other models, followed closely by \system{VisionSpeechCNN}.
Visual context is clearly beneficial in matching soft human ratings.
Next, we further analyse the models by addressing the following questions.

\myparagraph{Does \system{VisionSpeechCNN} only output common words?}
The 
baseline models (rows 1 and 2) respectively assign scores to keywords according to their occurrence in the training transcriptions and the average visual tagger output.
\system{VisionSpeechCNN} outperforms both across all metrics.

\myparagraph{Does the model do more than map acoustics to images?}
One possibility is that \system{VisionSpeechCNN} 
might 
learn to in effect ignore signals in the acoustics that are irrelevant to producing the visual output.
To see if this is so, we compare with \system{VisionCNN} (row 4), which represents a test utterance by passing the paired test image through the visual tagger. 
\system{VisionSpeechCNN} outperforms \system{VisionCNN} across all metrics, except for Spearman's $\rho$.
This indicates that \system{VisionSpeechCNN} does not simply reproduce the output from the visual tagger (used to supervise it); it actually achieves superior performance for all exact keyword spotting metrics and most of the semantic metrics.
Again, Spearman's $\rho$ takes the actual annotator counts into account, and here \system{VisionCNN} performs best.

\myparagraph{Is there a benefit to using visual context over transcriptions?}
This work is motivated by settings in which text transcriptions are not available while images may be.  However, it is interesting 
to consider whether there is some benefit to the training images beyond serving as weak labels.
Might the visual grounding actually provide better supervision than text for some purposes?

\begin{figure*}[t]
    \centering
    \includegraphics[width=0.48\linewidth]{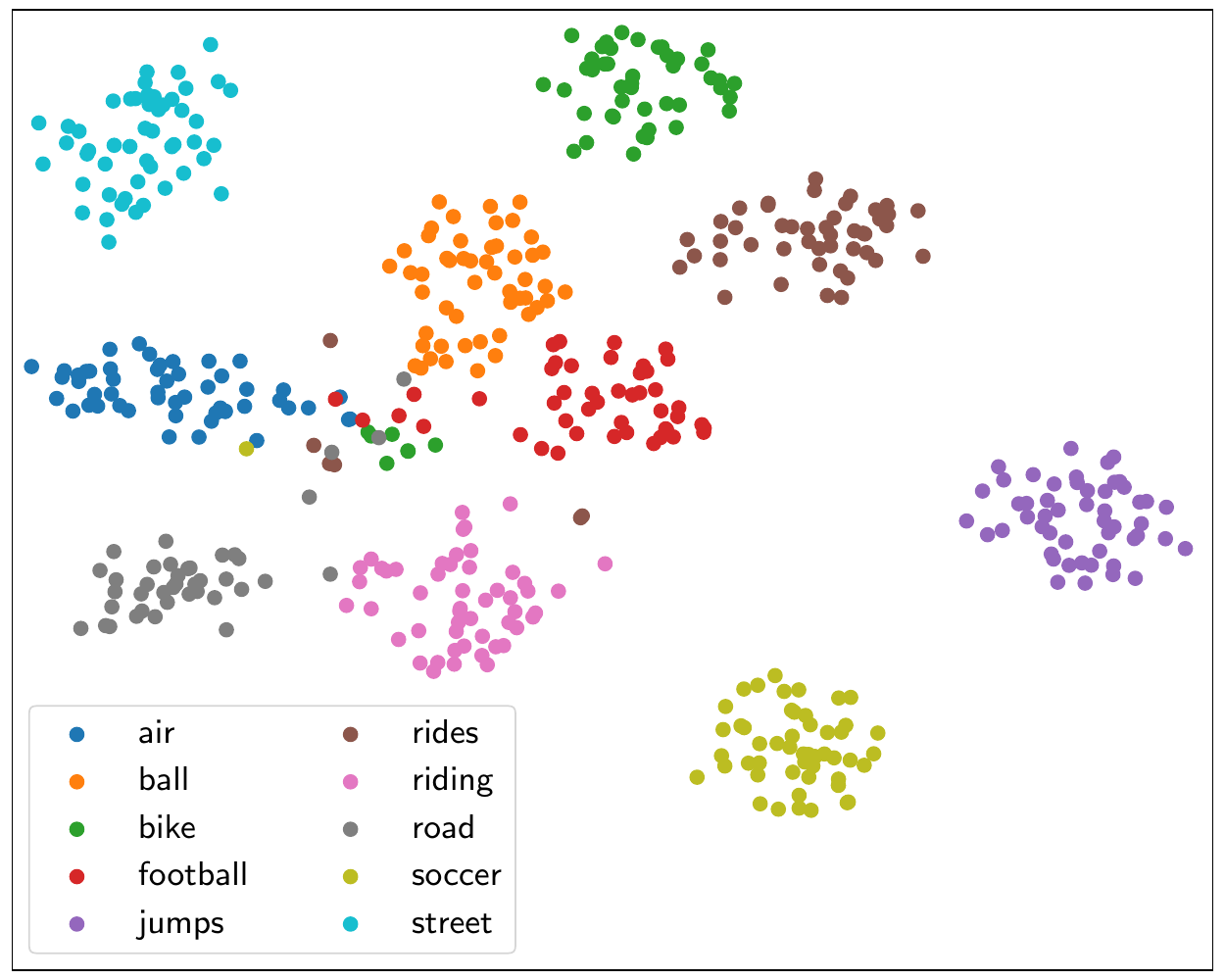}\hfill\includegraphics[width=0.48\linewidth]{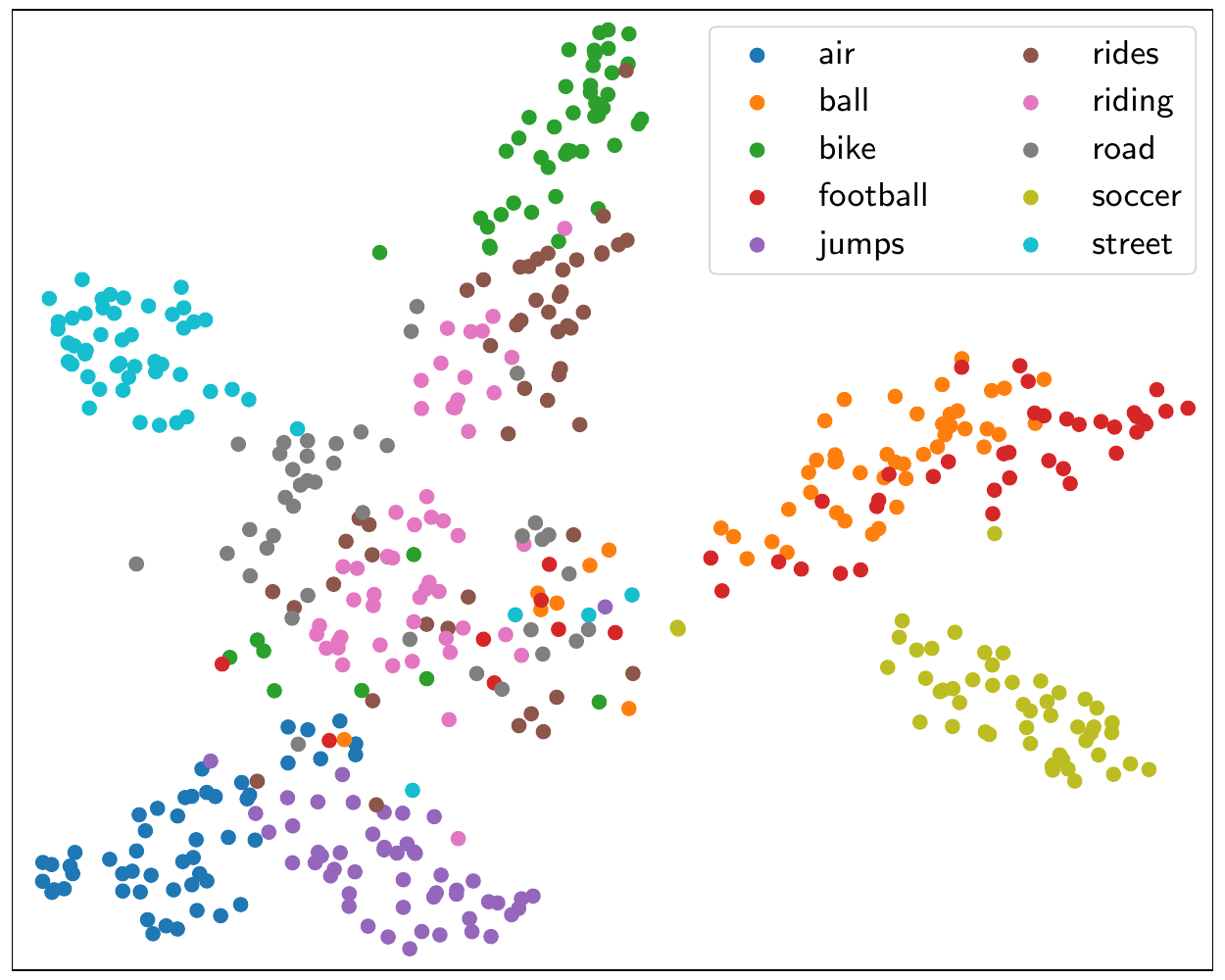}\\
    \vspace*{-2.5pt}
    {\footnotesize ~\hfill (a) \textsc{SupervisedBoWCNN} \hfill~~\hfill (b) \textsc{VisionSpeechCNN} \hfill~ }
    \vspace*{-5pt}
    \caption{t-SNE visualisations of acoustic embeddings of isolated words for ten keyword types.
    The embeddings were obtained from penultimate bottleneck layers for words from the development data. 
    }
    \label{fig:tsne}
\end{figure*} 

We compare rows 3 and 5 of Table~\ref{tbl:results}.
In keyword spotting, \system{VisionSpeechCNN} lags behind \system{SupervisedBoWCNN} (trained on ground truth transcriptions). 
{However, when moving to semantic speech retrieval, \system{VisionSpeechCNN} improves on almost all metrics, while \system{SupervisedBoWCNN} performs worse on most scores relative to exact keyword spotting.}
\system{VisionSpeechCNN} also outperforms \system{SupervisedBoWCNN} on Spearman's $\rho$, so it matches better the soft human judgements.  
This indicates that visual grounding actually provides better supervision than text for some purposes.

Despite the benefit of visual supervision, \system{SupervisedBoWCNN} still performs better on semantic speech retrieval measured 
against hard labels (in absolute scores). 
{We now show that this is because many exact matches are also semantic matches, which \system{SupervisedBoWCNN} predicts well, while \system{VisionSpeechCNN} actually performs better on non-verbatim semantic matches.
To do this, we consider separately the contributions of exact and exclusively semantic matches for the semantic $P@N$ metric (which actually also captures recall, since $N$ is the number of true occurrences of a keyword). Standard}
$P@N$ cannot be broken into separate components, 
since it is averaged over keywords.
We therefore define an unweighted $P@N^* = \frac{\sum_w c_w}{\sum_w N_w}$, where $N_w$ is the number of occurrences of keyword $w$ in the ground truth labels, and $c_w$ is the number of correct utterances in the top $N_w$ predictions. 
Count $c_w$ can then be broken into exact and semantic matches, such that  $P@N^* = P@N^*_{\textrm{exact}} + P@N^*_{\textrm{sem}}$ (here $P@N^*_{\textrm{sem}}$ considers only non-exact semantic matches).
Table~\ref{tbl:p_at_n_breakdown} shows that the $P@N^*$ of \system{SupervisedBoWCNN} is dominated by correct exact predictions, with a contribution of 11.7\% from semantic matches.
In contrast, \system{VisionSpeechCNN} makes more correct semantic (25.3\%) than exact (22.3\%) predictions.
The visual model \system{VisionCNN} has the highest proportion of semantic matches by a small margin.  The non-exact match results track well the Spearman's $\rho$ results in Table~\ref{tbl:results}. 

\begin{table}[t]
    \mytable
    \caption{Unweighted semantic 
    $P@N^* = P@N^*_{\textrm{exact}} + P@N^*_{\textrm{sem}}$ separates out the contributions from exact and exclusively semantic matches. Scores are given as percentages (\%).}
    \begin{tabularx}{\linewidth}{@{}lCCC@{}}
    \toprule
    Model & $P@N^*$ & $P@N^*_{\textrm{exact}}$ & $P@N^*_{\textrm{sem}}$ \\
    \midrule
    \textsc{VisionSpeechCNN} & 47.5 & 22.3 & 25.3 \\
    \textsc{VisionCNN} & 44.7 & 19.2 & 25.5 \\
    \textsc{SupervisedBoWCNN} & 50.0 & 38.4 & 11.7 \\
    \bottomrule
    \end{tabularx}
    \label{tbl:p_at_n_breakdown}
\end{table}

For a qualitative comparison of 
\system{VisionSpeech\-CNN} and \system{SupervisedBoWCNN}, we passed a set of isolated segmented spoken words through
both models.
Figure~\ref{fig:tsne} shows t-SNE embeddings \cite{vandermaaten+hinton_jmlr08} of representations from a 256-dimensional bottleneck layer (used for computational reasons) added between the 3000-dimensional ReLU layer and the final output of both models.
Although words are well-separated in the case of the transcription-supervised model~(a), the visually grounded model are more successful in mapping semantically related spoken words to similar embeddings~(b): related spoken words like `bike', `rides' and `riding' have similar embeddings, as do `air' and `jumps', and `football' and `ball' (which is also closer to `soccer').

\myparagraph{What is the best we could do {with a cascaded approach}?}
The cheating models \system{TextWuP} and \system{TextParagram} (rows 6 and 7, Table~\ref{tbl:results}) represent the setting where we have access to a perfect ASR system. 
On the metrics that use hard ground truth labels, these two models perform best.
However,
\system{VisionCNN} and \system{VisionSpeechCNN} 
perform better on
Spearman's $\rho$. 
As noted, the visually trained models are particularly strong in matching non-exact semantic keywords.

\myparagraph{How does cascaded retrieval perform in low-resource settings in comparison to visual grounding?}
The previous question above considered the setting where perfect ASR is available.
Our main focus is on extreme low-resource settings where no ASR is available. 
But in some low-resource settings, an intermediate situation may exist where limited amounts of data might still be available for ASR development.
It is therefore interesting to consider how a cascaded model such as \system{TextParagram} would perform with actual ASR and how performance changes as ASR accuracy deteriorates.
To simulate systematically worse low-resource settings, we artificially introduce errors in the ground truth transcriptions,\footnote{Given a target word error rate $e$, for each word in the ground truth transcriptions, we introduce an error to it with probability $e$ and then uniformly pick a substitution, deletion or insertion.
} and then apply \system{TextParagram} to this simulated ASR output.
We also use an actual ASR system (Google Cloud Speech-to-Text\footnote{\url{https://cloud.google.com/speech-to-text/}}), but this is only a reference since in typical low-resource settings such an accurate system would not be available.

Semantic speech retrieval results are shown in Table~\ref{tbl:cascaded_asr}, with \system{SimASR} indicating the cascaded approach with the simulated ASR systems.
Performance across all metrics systematically deteriorates as ASR accuracy becomes worse.
The cascaded approach with \system{SimASR} at a word error rate (WER) of 50\% performs similarly to \system{VisionSpeechCNN} on $P@10$ and $P@N$
(bottom row), but worse on all other metrics.
This indicates that \system{VisionSpeechCNN} could prove useful even when ASR is available (with a relatively low accuracy).
However, further experiments on truly low-resource languages are required to illustrate this more conclusively.

\begin{table}[t]
    \mytable
    \caption{Semantic speech retrieval performance (\%) using a cascaded approach combining ASR with \textsc{TextParagram}.  The word error rate of each ASR system is given in parentheses.  \textsc{PerfectASR} reproduces the \textsc{TextParagram} results (last row) of Table~\ref{tbl:results}.  \textsc{GoogleASR} uses an actual high-resource ASR system cascaded with \textsc{TextParagram}.  \textsc{SimASR} uses a simulated ASR system with varying error rates cascaded with \textsc{TextParagram}.}
    \begin{tabularx}{\linewidth}{@{}lCCCCc@{}}
    \toprule
    Method (WER) & $P@10$ & $P@N$ & EER & AP & Spear.\ $\rho$ \\
    \midrule
    \textsc{PerfectASR} (0\%) & \textbf{88.8} & \textbf{64.0} & \textbf{14.3} & 60.1 & 31.6 \\
    \textsc{GoogleASR} (8.6\%) & 88.2 & 62.8 & 15.5 & 58.8 & 30.8 \\
    \textsc{SimASR} (8.6\%) & 85.5 & 61.0 & 16.7 & 56.0 & 27.7 \\
    \textsc{SimASR} (20\%) & 77.5 & 55.3 & 20.7 & 49.1 & 26.8 \\
    \textsc{SimASR} (50\%) & 60.5 & 40.9 & 30.9 & 31.4 & 19.2 \\
    \textsc{SimASR} (80\%) & 34.6 & 22.3 & 40.8 & 15.0 & \hphantom{0}9.6 \\
    \addlinespace
    \textsc{VisionSpeechCNN} & 58.8 & 39.7 & 23.9 & 39.4 & \textbf{32.4} \\
    \bottomrule
    \end{tabularx}
    \label{tbl:cascaded_asr}
\end{table}

\myparagraph{Are the text-based models an alternative to human judgements?}
Since the image-speech corpus has transcriptions, 
we may ask whether it is necessary to collect human annotations at all, or whether they can be generated automatically 
by a semantic model like our \system{Text} cheating models.
Although the \system{Text} methods match hard human labels better than other approaches in Table~\ref{tbl:results}, they are far from perfect.
Furthermore, when evaluated against the soft annotator counts, \system{VisionCNN} and \system{VisionSpeechCNN} actually do better, indicating that the \system{Text} models cannot replace human judgements.

%% file: conclusion.tex
\section{Conclusions and Future Work}

We investigated how a model that learns from parallel images and unlabelled speech captures aspects of semantics in speech.
We collected a new data set for a
semantic speech retrieval task, where the aim is to retrieve utterances that are semantically relevant given a written query keyword.
Without seeing any parallel speech and text, the
visually grounded model achieves a semantic $P@10$ of almost 60\%.
Although a model trained on transcriptions is superior on some metrics, the vision-speech model retrieves more than double the number of non-verbatim semantic matches and is a better predictor of the actual soft human ratings.  

Since visual context seems to provide different information from transcriptions for semantic speech retrieval, future work could consider how these two supervision signals could be combined when both are available. 
We would also like to explore whether our model could be used to localise the parts of the input signal giving rise to a particular (semantic) label, similarly to analyses in studies that inspired this work~\cite{palaz+etal_interspeech16,harwath+glass_acl17}.
Finally, many cognitively motivated studies have proposed to combine spoken and visual input to model infant language acquisition, but few use real unlabelled speech with natural visual input as we do here.
The cognitive implications of our model 
could be a topic for future exploration. 

We encourage interested readers to make use of our data and associated task, 
which we make available at: {\url{https://github.com/kamperh/semantic_flickraudio/}}.

\section*{Acknowledgements}
We would like to thank Shane Settle, John Wieting, and Michael Roth for helpful discussions and input.
